\documentclass[letterpaper]{article} 
\usepackage{aaai2026}  
\usepackage{times}  
\usepackage{helvet}  
\usepackage{courier}  
\usepackage[hyphens]{url}  
\usepackage{graphicx} 
\usepackage{natbib}  
\usepackage{caption} 
\usepackage{algorithm}
\usepackage{algorithmic}

\usepackage{newfloat}
\usepackage{listings}

\usepackage{amsmath}
\usepackage{amsfonts}

\usepackage{booktabs}
\usepackage{multirow}

\usepackage{xcolor}

\DeclareCaptionStyle{ruled}{labelfont=normalfont,labelsep=colon,strut=off} 
\floatstyle{ruled}
\newfloat{listing}{tb}{lst}{}
\floatname{listing}{Listing}
\title{On the Fast Adaptation of Delayed Clients in Decentralized Federated Learning: A Centroid-Aligned Distillation Approach}
\author{
    Jiahui Bai\textsuperscript{\rm 1},  
    Hai Dong\textsuperscript{\rm 1},  
    A. K. Qin\textsuperscript{\rm 2}  
}
\affiliations{
    \textsuperscript{\rm 1}School of Computer Technologies, RMIT University, Melbourne, VIC, Australia\\
    \textsuperscript{\rm 2}School of Science, Computing and Engineering Technologies, Swinburne University of Technology, Hawthorn, VIC, Australia\\
    s3735049@student.rmit.edu.au, hai.dong@rmit.edu.au, kqin@swin.edu.au
}

\usepackage{bibentry}

\begin{document}

\maketitle

\begin{abstract}
Decentralized Federated Learning (DFL) struggles with the slow adaptation of late-joining delayed clients and high communication costs in asynchronous environments. These limitations significantly hinder overall performance. To address this, we propose DFedCAD, a novel framework for rapid adaptation via Centroid-Aligned Distillation. DFedCAD first employs WCP to compress models into representative centroids, drastically reducing communication overhead. It then enables delayed clients 
to compute weighted alignment 
with peer knowledge using a novel structural distance metric and a differentiable $k$-means distillation module, facilitating efficient 
knowledge transfer. Extensive experiments on CIFAR-10, CIFAR-100, and Tiny-ImageNet show that DFedCAD consistently achieves state-of-the-art performance, attaining the highest accuracy across all evaluated settings while reducing communication overhead by over 86\% and computational cost by 42\%. Our framework provides a 
practical solution for efficient decentralized learning in dynamic
and delay-prone environments.
\end{abstract}


\section{Introduction}
The explosive increase in distributed data, combined with escalating concerns over data privacy, has positioned Federated Learning (FL) as an attractive approach for collaborative machine learning \cite{mcmahan2017communication}. By allowing multiple participants to jointly train a model without directly sharing their original data, Federated Learning effectively protects user privacy and ensures adherence to regulatory standards \cite{9599369}. However, mainstream FL approaches often rely on a central server to coordinate model aggregation, which introduces issues such as single points of failure, bottlenecks, and trust management \cite{dai2022dispfl}. To overcome these limitations, Decentralized Federated Learning (DFL) has garnered increasing attention. DFL eliminates the need for a central coordinator by enabling peer-to-peer model exchanges, thereby enhancing system robustness and privacy \cite{9850408}. 

Despite its advantages, DFL introduces new challenges in practice. In particular, its reliance on peer-to-peer communication makes it more sensitive to issues such as \textit{asynchronous participation} and \textit{delayed client activation}. These issues can lead to slower convergence and higher communication overhead, especially in the absence of global synchronizations
\cite{gholami2023fast, bornstein2022swift}. Among these, the presence of delayed clients poses a particularly critical challenge for decentralized systems. In this context, a \textit{delayed client} refers to one that joins training after it has already begun, rather than participating from the initial training rounds. 
Unlike classical asynchronous FL, where clients may update at different times but are generally present throughout the training process, delayed clients entirely miss earlier model evolution. As a result, they often lack historical model context, making effective adaptation particularly challenging in decentralized settings.

Although many studies have explored communication-efficient approaches in federated learning, most are built upon centralized and synchronous aggregation paradigms \cite{gupta2022survey, yuan2024decentralized}. These methods are often ill-suited for decentralized settings, where communication is peer-to-peer and client participation may be delayed or asynchronous. In particular, newly joined clients typically lack effective guidance. At the same time, communication overhead remains a critical bottleneck for 
training efficiency. These challenges underscore the need for new mechanisms that can accelerate client adaptation while maintaining communication efficiency in decentralized environments.

To achieve efficient communication and rapid adaptation in decentralized federated learning, it is essential to design mechanisms that enable newly joined clients to effectively leverage the knowledge accumulated by their neighbors, even under constrained communication and computation budgets. However, such mechanisms remain underexplored, limiting the 
practical applicability of existing decentralized systems. This leads to our central research question:

\begin{center}
\textit{How can we design decentralized FL systems that support rapid adaptation of delayed clients under limited communication and computation budgets?}
\end{center}

To address these challenges, we propose DFedCAD, a novel decentralized federated learning framework designed to enable rapid adaptation of delayed clients while reducing communication and computation costs. At its core, DFedCAD employs a WCP mechanism. Each client applies a pruned $k$-means algorithm to the weights of every layer, grouping them into $K$ clusters, and exchanges only the $K$ centroids along with their index sequences with their neighbors. This strategy compresses full-precision parameter tensors into a handful of representative centroids, thereby reducing transmission overhead while preserving model expressiveness.


To facilitate fast adaptation, DFedCAD introduces a collaborative alignment strategy. Delayed clients collect centroids from their neighbors and compute relevance scores based on centroid distribution distance. These scores are then used to construct a weighted multi-teacher guidance signal. A differentiable $k$-means layer is employed to softly align the local model with this multi-source reference under teacher supervision.

In addition, DFedCAD incorporates a personalization-guided training mechanism. By measuring the difference between the current model and a reference model, the client adjusts its gradient updates to balance between adaptation and personalization. This helps stabilize optimization and accelerates convergence for newly joined clients. Through the integration of lightweight communication, collaborative alignment, and structured training, DFedCAD provides an effective solution for decentralized federated learning.

\noindent Our main contributions are summarized as follows:
\begin{itemize}
    \item 
    We propose DFedCAD, the first framework that combines WCP with multi-teacher alignment to enable rapid adaptation of newly joined clients under limited communication and computation budgets.

    \item 
    We develop a centroid distribution distance metric to assess the relevance of neighbor models, and construct a multi-teacher guidance signal using a differentiable $k$-means layer, which enables delayed clients to softly align their local models with peer knowledge.

    \item 
    Extensive experiments on CIFAR-10, CIFAR-100, and Tiny-ImageNet show that DFedCAD consistently 
    outperforms existing baselines in accuracy
    across all datasets, while significantly reducing communication and computation overhead compared to state-of-the-art decentralized FL methods.
\end{itemize}

\section{Related Work}
\subsection{Communication- and Computation-Efficient Federated Learning}
Improving communication and computational efficiency remains a fundamental challenge in FL, especially for resource-constrained edge devices with limited bandwidth and high latency. Existing research has proposed various methods to address these challenges.
Approaches like FedBiOAcc \cite{li2023communication} and AggITD \cite{xiao2023communication} achieve reduced communication overhead through variance-reduced gradient estimation. DoCoFL \cite{dorfman2023docofl} and QFedCG \cite{xu2023federated} further enhance communication efficiency by employing gradient sparsification and quantization techniques, with QFedCG uniquely adapting compression levels based on individual client capabilities. However, these methods typically rely on centralized aggregation and synchronous communication, limiting their effectiveness in DFL scenarios.

Recent decentralized approaches, including SWIFT \cite{bornstein2022swift}, and DFedPGP \cite{liu2024decentralized}, utilize gossip-based communication and partial gradient exchange to improve model consistency and personalization. Despite these advances, they predominantly assume continuous, synchronized client participation, thereby inadequately addressing challenges related to delayed or intermittent client availability.

To thoroughly evaluate the performance of methods designed for decentralized and asynchronous environments, we select DFedPGP, representing decentralized approaches utilizing partial gradient exchanges, and QFedCG, exemplifying centralized gradient compression strategies, as representative baselines. These choices enable a comprehensive assessment of the proposed framework under realistic decentralized conditions involving delayed participation and constrained communication resources.

\subsection{Knowledge Distillation in Federated Learning}
Knowledge distillation (KD) has emerged as a key technique in federated learning (FL) to address data and model heterogeneity. KD transfers knowledge via model outputs or intermediate representations, providing a lightweight alternative to parameter aggregation, especially suited for privacy-sensitive settings.

Centralized KD methods such as FedFed \cite{yang2023fedfed} and FedHKD \cite{chen2023best} aggregate predictions centrally, effectively handling non-IID data. FedIOD \cite{gong2024federated} and FedSD2C \cite{zhangone} further eliminate reliance on real public datasets through synthetic distillation,
where clients collaboratively generate auxiliary data using generative models and then distill knowledge based on these synthetic samples.
However, these methods inherently require centralized servers and global synchronization, limiting their utility in decentralized and asynchronous environments.

Approaches addressing model and data heterogeneity include ReT-FHD \cite{qirethinking}, which adapts temperature scaling for heterogeneous models, and Fed-DFA \cite{wang2025fed} and DFRD \cite{wang2023dfrd}, which apply adversarial and generative distillation, respectively. FedGMKD \cite{zhang2024fedgmkd} and Spectral Co-Distillation \cite{chen2023spectral} further enhance personalization through prototype-driven and spectrum-aware methods. Nevertheless, their dependence on synchronous updates and structured global coordination restricts applicability to environments with delayed or intermittent client participation.

Recent decentralized KD methods like DESA \cite{huang2024overcoming} and multi-headed distillation \cite{zhmoginov2023decentralized} avoid central aggregation by leveraging synthetic data or peer-to-peer interactions. However, these methods still rely on public data, synchronization, or intensive computation, limiting their effectiveness in dynamic and resource-constrained decentralized environments.

To address these limitations, our work focuses on decentralized FL with asynchronous participation, delayed clients, and limited communication. We adapt ReT-FHD \cite{qirethinking} and MTKD-RL \cite{yang2025multi}—the latter being state-of-the-art in multi-teacher KD that uses reinforcement learning for adaptive weighting—as baselines. Our approach employs centroid-based collaborative alignment, enabling efficient peer-to-peer knowledge transfer without requiring full model synchronization or centralized coordination.

\section{Problem Formulation}
We consider decentralized federated learning (DFL), where each client collaborates with its neighbors to learn a client-specific model under limited communication and computation budgets. Let $\mathcal{C}$ denote the set of 
delayed
clients. Each client $i \in \mathcal{C}$ holds a private dataset $S_i$ sampled from an unknown local distribution $D_i$ over the input-label space $X \times Y$. Due to statistical heterogeneity among clients, it is more appropriate to learn a local model $h_{\theta_i} \in \mathcal{H}: X \rightarrow Y$ for each client $i$, rather than enforcing a global model shared across all clients, which may be suboptimal under statistical heterogeneity.

To enable personalization while promoting knowledge sharing from neighboring clients, we formulate the following objective:
\begin{equation}
\min_{\{\theta_i\}_{i \in \mathcal{C}}} \sum_{i \in \mathcal{C}} \mathbb{E}_{(x,y)\sim D_i} \left[\ell(h_{\theta_i}(x), y) + \alpha \cdot \mathcal{L}_{\text{align}}(\theta_i, \{\tilde{\theta}_j\}_{j \in \mathcal{N}_i})\right]
\end{equation}

Here, $\ell: X \times Y \rightarrow \mathbb{R}^+$ is a standard supervised loss (e.g., cross-entropy), and $\mathcal{L}_{\text{align}}$ is an alignment loss that distills knowledge from the neighbor 
compression
models $\{\tilde{\theta}_j\}_{j \in \mathcal{N}_i}$ into the local model $\theta_i$. The neighbor set $\mathcal{N}_i$ corresponds to the peers with which client $i$ can communicate in a decentralized topology. The alignment loss 
distills knowledge from 
compressed model representations, as detailed in Section~\ref{sec:dfm_align}, to align the local model with those of neighboring clients. The hyperparameter $\alpha \geq 0$ controls the trade-off between local fitting and knowledge alignment.

The first term ensures that each client model fits its own local data, while the second term guides delayed clients to align with the compressed knowledge shared by their peers, enabling faster adaptation and communication-efficient collaboration.

\section{Decentralized Federated Learning via Centroid-Aligned Distillation}
\begin{algorithm}[t]
\caption{DFedCAD Decentralized Training}
\label{alg:dfedcad}
\begin{algorithmic}[1]

\REQUIRE 
  clients $C$ with local datasets $\mathcal{D}$, join rounds $\{\tau_i\}$, total rounds $R$, batch size $B$, learning rate $\eta$, number of peers $n$
\STATE $S \gets \{\,i\in C\mid \tau_i = 0\}$

\FORALL{$i \in S$}
\STATE Initialize local model $\theta_i^0$
\ENDFOR
\FOR{$r = 1 \to R$}
\STATE $N \gets \{\,i\mid \tau_i = r\}$
\FORALL{$i \in N$ {\bf parallel}}
\STATE Initialize local model $\theta_i^0$
\STATE $\mathcal{B}$ $\gets$ \text{(split datasets $\mathcal{D}_i$ into batches of size $B$)}
\STATE Update $\theta_i^{r} \gets \theta_i^0 - \eta\nabla\ell(\theta_i^0; b)$ for each $b \in \mathcal{B}$
\ENDFOR
\STATE $S \gets S \cup N$
\STATE Build peer graph $G^r$ where each $i \in S$ randomly selects $n$ peers
\FORALL{$i \in S$ {\bf parallel}}
\STATE $\theta_i^{r+1} \gets \texttt{LocalUpdate}(\{\tilde\theta_j^r: G^{r-1}_{ji}=1\}, M_i^{r})$
\FOR{each layer $\ell$ in $\theta_i^{r+1}$}
    \STATE $\tilde{\theta}_{i,\ell}^{r+1}, M_{i,\ell}^{r+1} \gets \texttt{WCP}(\theta_{i,\ell}^{r+1})$
\ENDFOR
\STATE Send $\tilde{\theta}_i^{r+1}$ to all neighbors $j$ with $G^r_{ij} = 1$
\ENDFOR
\ENDFOR
\end{algorithmic}
\end{algorithm}

\subsection{Overview of DFedCAD}
The overall training workflow of DFedCAD is illustrated in Algorithm 1 and Algorithm 1 of Appendix A.1, which together define a decentralized federated learning process. In the first round ($r = 0$), all clients with $\tau_i = 0$ participate in training, forming the initial client set $S$ (line 1). Each client independently initializes its local model parameters $\theta_i^0$ (lines 2–4). In every subsequent round $r$, newly 
delayed 
clients $N$ (line 6) join the training process by performing a lightweight warm-up procedure on their local datasets to initialize their local models (lines 7–10), allowing them to integrate smoothly into the current collaboration state.
This warm-up ensures that the local model captures basic data characteristics, enabling a meaningful comparison between its feature-space centroid and that of the teacher model during alignment.

In Algorithm 1 of Appendix A.1, each client first computes the average of received neighbor compressed models to obtain a reference model $\theta_{ref}$ (lines 1–2), which serves as a direction for momentum-based updates. This momentum strategy helps achieve knowledge sharing across clients.
The client then splits its local dataset into batches (line 3) and determines whether structural alignment is required based on its role. For delayed clients, importance weights ${\alpha_j}$ are computed from received models using formulas~\ref{equ:cdf}–\ref{equ:importance} (lines 4–6), guiding the subsequent alignment process.

During the local training phase (lines 7–16), each client performs $E$ epochs of gradient descent. For every mini-batch, the masked supervised loss $L_{\mathrm{sup}}$ is computed using the current sparse model (line 9). If the client is delayed, it further invokes the DKM-Align module to perform structural alignment and obtains an alignment loss $L_{\mathrm{align}}$ (line 11), which is then combined with $L_{\mathrm{sup}}$ to yield the total loss $L = L_{\mathrm{sup}} + \lambda L_{\mathrm{align}}$ (line 12). The final model update is performed using the gradient of this total loss, with an additional momentum term $\gamma(\theta - \theta_{ref})$ based on the reference model (line 14), completing one round of local update. 

Overall, DFedCAD addresses the challenges posed by asynchronous client participation and heterogeneous model structures through integrating structural alignment and weight-cluster pruning. Notably, for delayed clients, DFedCAD significantly improves their ability to align with the structural patterns of peer models, enabling them to quickly catch up with ongoing training and effectively accelerate global convergence while mitigating performance degradation.

\subsection{Weight Clustering Pruning}
\label{sec:wcp}
To reduce communication overhead in decentralized federated learning, we adopt the Weight Clustering Pruning (WCP) method. The specific WCP algorithm can be found in Algorithm 2 of Appendix A.2. Specifically, WCP compresses model parameters layer-wise by performing weight clustering, where one centroid per layer is fixed at zero to automatically prune weights near zero, thereby achieving sparsity. 
Instead of transmitting full model parameters, clients only send a table of centroid values and the indices corresponding to which weights map to each centroid, significantly decreasing communication costs.

Formally, given a model layer with $N$ weights,
these weights are clustered into $k$ centroids ($k \ll N$) in partitions, the traditional communication overhead is $N \times B$ bits (with $B$ bits per weight). In contrast, WCP reduces this overhead to $k \times B$ bits for the centroid table plus $N \times \lceil \log_2 k \rceil$ bits for the index sequence. By iteratively updating non-zero centroids and mapping weights to their closest centroid, the zero centroids can dynamically adjust pruning intensity and model sparsity to generate sparse masks $M$. Further detailed analysis and visualization of WCP are available in Appendices A.2 and C. 

In this work, we further utilize the centroids generated from WCP as foundational representations for collaborative knowledge alignment. Specifically, the centroids not only serve as compressed parameters to reduce communication costs, but also enable the construction of multi-teacher guidance signals across clients, facilitating rapid adaptation of newly joined clients to the knowledge of their neighbors.

\subsection{Centroid Distribution Distance and Teacher Weighting}
\label{sec:cdf}

To enable delayed clients to effectively leverage knowledge accumulated by their neighbors, we propose a centroid-based distribution discrepancy measure and a corresponding teacher weighting strategy. Specifically, we first apply WCP to compress each client's model weights into a set of representative centroids. These centroids constitute a compact representation of each client's structural distribution, transforming the comparison of client models into the measurement of centroid distribution differences.

Inspired by recent advancements in distribution matching, we adopt the Characteristic Function Distance (CFD), which was initially designed for comparing data distributions ~\cite{wang2025dataset}. 
We extend CFD to the parameter space by treating the $k$ centroids of each model layer as an empirical distribution and computing the squared $L_2$ distance between their characteristic functions over a fixed set of frequency vectors. 
This centroid-based CFD quantifies structural discrepancies between local and neighbor models.

Formally, the characteristic function (CF) of a centroid set $\mathcal{M} = \{\mu_1, \dots, \mu_k\}$, viewed as a discrete uniform distribution in parameter space, is defined as: 
\begin{equation}
\Phi_\mu(t) = \frac{1}{k} \sum_{i=1}^{k} e^{j\langle t, \mu_i \rangle},
\end{equation}
where $t \in \mathbb{R}^d$ is a frequency vector sampled from a Gaussian distribution.  
Given two clients with centroid sets $\mathcal{M}^{(i)}$ and $\mathcal{M}^{(j)}$, the Characteristic Function Distance (CFD) between them is computed as:

\begin{equation}
\label{equ:cdf}
\text{CFD}(\mathcal{M}^{(i)}, \mathcal{M}^{(j)}) = \mathbb{E}_{t \sim \mathcal{N}(0, \sigma^2 I)} \left[\left|\Phi_{\mathcal{M}^{(i)}}(t) - \Phi_{\mathcal{M}^{(j)}}(t)\right|^2\right],
\end{equation}
where the expectation is approximated by Monte Carlo averaging over $n$ sampled frequencies.

In practice, we first compute the CFD for each clustered layer independently and subsequently average these layer-wise CFD values, yielding a single structural discrepancy measure between two client models. After obtaining CFD values for all neighbor clients, we apply min-max normalization to these scores as follows:
\begin{equation}
\label{equ:score}
\hat{s}_j = \frac{\text{CFD}_j - \min_{k}\text{CFD}_k}{\max_{k}\text{CFD}_k - \min_{k}\text{CFD}_k + \varepsilon},
\end{equation}
where $\varepsilon$ is a small constant introduced to ensure numerical stability. Finally, we compute the teacher importance weights $\alpha_j$ using a softmax function applied to these normalized CFD values:
\begin{equation}
\label{equ:importance}
\alpha_j = \frac{\exp(-\hat{s}_j)}{\sum_{j'}\exp(-\hat{s}_{j'})},
\end{equation}
These computed teacher importance weights guide the delayed client's training through multi-teacher alignment, assigning greater importance to teacher models with smaller structural discrepancies, thus enabling efficient and rapid knowledge transfer from the most structurally relevant neighbors.

\subsection{Differentiable $K$-Means Alignment Module}
\label{sec:dfm_align}
To facilitate efficient structural alignment between delayed client models and their neighbor (teacher) models, we propose a differentiable $k$-means (DKM) alignment module. Traditional differentiable $k$-means clustering methods have primarily been used for parameter compression or reconstruction ~\cite{cho2021dkm}. We further extend this approach to achieve structural alignment across client models. Specifically, we leverage a centroid set as structural anchors, while the soft assignment matrix between model weights and centroids encodes the structural distribution pattern of parameters within each layer. 
The specific DKM algorithm can be found in Algorithm 3 of Appendix A.3.

Formally, given a student's parameter matrix $W \in \mathbb{R}^{N \times D}$, where $N$ denotes the number of weights and $D$ is the parameter dimensionality, our DKM module iteratively updates clustering through differentiable Expectation-Maximization steps as follows:

In the E-step, we first calculate the squared Euclidean distances between each weight and centroid. We then convert these distances into a soft assignment matrix $A_S \in \mathbb{R}^{N \times K}$ via a softmax operation:
\begin{equation}
\label{euq:soft_ass_mat}
A_{S,n,k} = \frac{\exp(-\|W_n - C_k\|_2^2)}{\sum_{k'=1}^{K}\exp(-\|W_n - C_{k'}\|_2^2)},
\end{equation}
where $C_k$ represents the $k$-th centroid.

In the M-step, centroids are updated according to the current soft assignments:
\begin{equation}
\label{euq:soft_ass}
C_k \leftarrow \frac{\sum_{n=1}^{N} A_{S,n,k} W_n}{\sum_{n=1}^{N} A_{S,n,k} + \varepsilon},
\end{equation}
where $\varepsilon$ is a small constant ensuring numerical stability. Similarly, each neighbor (teacher) client model provides its own centroid sets $\{C^{(t)}_k\}_{k=1}^{K}$ and corresponding soft assignment matrices $A^{(t)}_T \in \mathbb{R}^{N\times K}$. To structurally align the student's model with its neighbors, we introduce a hybrid similarity metric that combines semantic and numerical similarities. First, we define the semantic similarity as the Jaccard similarity between soft assignments from the student and teacher models:
\begin{equation}
\label{euq:j_sim}
J^{(t)}_{i,j} = \frac{\sum_{n=1}^{N}\min(A^{(t)}_{T,n,i}, A_{S,n,j})}{\sum_{n=1}^{N}\max(A^{(t)}_{T,n,i}, A_{S,n,j})+\varepsilon}.
\end{equation}
This captures how closely student and teacher centroids cover similar subsets of model weights. Next, the numerical similarity between centroids from the teacher and student is defined based on Euclidean distance:
\begin{equation}
\label{euq:e_distan}
S^{(t)}_{i,j} = \exp\left(-\beta_{\text{dist}}\|C_i^{(t)} - C_j\|_2^2\right),
\end{equation}
where the parameter $\beta_{\text{dist}}>0$ modulates sensitivity to centroid distances. In the experiment $\beta_{\text{dist}}$ is  set to 1.0. We then integrate semantic and numerical similarities via a weighted geometric mean to form the final matching weight matrix:
\begin{equation}
\label{euq:match_mat}
M^{(t)}_{i,j} = \left(J^{(t)}_{i,j}+\varepsilon\right)^{\alpha_{\text{mix}}} \left(S^{(t)}_{i,j}+\varepsilon\right)^{1-\alpha_{\text{mix}}},
\end{equation}
where $\alpha_{\text{mix}}\in[0,1]$ controls the relative emphasis between semantic and numerical similarity. In the experiment, $\alpha_{\text{mix}}$ is set to 0.5. The final matching weight for each teacher $t$ is obtained by row-wise normalization:
\begin{equation}
\label{euq:match_weight}
w^{(t)}_{i,j} = \frac{M^{(t)}_{i,j}}{\sum_{j'=1}^{K} M^{(t)}_{i,j'}+\varepsilon}.
\end{equation} 

Aggregating information from all teachers with the previously computed teacher importance weights $\{\alpha_t\}$, we construct teacher-aligned target centroids as follows:
\begin{equation}
\label{euq:tar_cen}
\widetilde C_{j} = \sum_{t=1}^{T}\alpha_t\sum_{i=1}^{K} w^{(t)}_{i,j} C_i^{(t)}.
\end{equation}

During training, we enforce structural alignment by minimizing a reconstruction loss with respect to the teacher-aligned centroids:
\begin{equation}
\label{euq:ali_loss}
\mathcal{L}_{\text{align}} = \frac{1}{N} \|W - A_S \widetilde C\|_F^2.
\end{equation}
The term $\| \cdot \|_F$ denotes the Frobenius norm, which computes the sum of squared differences across all matrix elements. Dividing by $N$, where $N$ is the number of weights, converts the reconstruction error into a Mean Squared Error (MSE), ensuring scale-invariant comparisons across layers of different sizes. $A_S \widetilde{C}$ represents a low-rank reconstruction of the student parameters $W$ using the teacher-aligned centroids $\widetilde{C}$. The alignment loss thus captures how well the teacher's structural basis can explain the student's parameters. By minimizing this error, the student is explicitly encouraged to align its internal structure with that of its neighbors.

Notably, this alignment loss $\mathcal{L}_{\text{align}}$ is fully differentiable, enabling the student centroids to move towards the teacher centroids during training, while simultaneously backpropagating gradients through the upper-level parameters. This achieves genuine 
structural knowledge transfer and optimization. 
For clarity and completeness, the full pseudo-code of the proposed DKM-Align algorithm, along with a detailed explanation of each step, is included in Algorithm 4 of Appendix A.3.

\subsection{Local Training with Centroid-Aligned Distillation}
\label{sec:alignment}
The local training phase in DFedCAD is crucial for enabling structural knowledge transfer and accelerating the adaptation of delayed clients. At the beginning of each communication round, delayed clients receive WCP-compressed centroid sets from their neighbors. These centroids provide both communication efficiency and a compact representation of peer model structures. To evaluate their relevance, clients compute the CFD between local and neighbor centroids in the frequency domain. The resulting distances are converted into teacher weights using a softmax, allowing the client to emphasize structurally similar models during alignment.

Training then proceeds on a sparsified version of the local model, guided by a pruning mask derived from WCP. During batch updates, the client optimizes a combined loss: a supervised classification loss and a structure-level distillation loss computed via the DKM module. This distillation process directly aligns the local parameters with the weighted teacher centroids and propagates alignment signals through differentiable gradients, shaping both model weights and their structural organization.

At the end of local training, the client re-applies WCP to update its centroids and pruning mask. This periodic refresh ensures that the compressed representation remains faithful to the evolving model structure, maintaining alignment quality in subsequent rounds.

Importantly, the three mechanisms are not standalone modules but interdependent components within a tightly integrated system. WCP defines the structural representation by compressing model parameters into centroids; CFD relies on these centroids to assess the distributional relevance of neighbor models; DKM, in turn, leverages both CFD-derived weights and WCP centroids to perform soft alignment. The gradients produced by DKM update the model parameters, which subsequently alter the centroids in the next WCP cycle. This feedback loop ensures that alignment, compression, and adaptation co-evolve synergistically, enabling stable and efficient learning in heterogeneous, decentralized settings.

\subsection{Convergence Analysis of DFedCAD}
\label{sec:convergence_dfedcad}

We analyze the convergence behavior of the DFedCAD algorithm under standard assumptions. All theoretical conditions and full proofs are deferred to Appendix E.

\paragraph{Assumption 1 (Smoothness of Local Objectives)}%
\label{assump:smooth}
Each local function $F_i(w)$ is $L$-smooth, i.e., $\|\nabla F_i(w) - \nabla F_i(v)\| \le L \|w - v\|$ for all $w,v\in\mathbb{R}^d$.

\paragraph{Assumption 2 (Strong Convexity)}%
\label{assump:strongly_convex}
Each $F_i$ is $\mu$-strongly convex with $\mu > 0$. The global objective $F(w) := \frac{1}{N} \sum_{i=1}^N F_i(w)$ has a unique minimizer $w^*$.

\paragraph{Assumption 3 (Stochastic Gradients)}%
\label{assump:bounded_variance}
The stochastic gradient $\nabla F_i(w; \xi)$ satisfies $\mathbb{E}[\nabla F_i(w; \xi)] = \nabla F_i(w)$ and $\mathbb{E}[\|\nabla F_i(w; \xi) - \nabla F_i(w)\|^2] \le \sigma_g^2$, with $\|\nabla F_i(w)\| \le G$.

\paragraph{Assumption 4 (Bounded Domain) \cite{zhang2022bort}}%
\label{assump:bounded_domain}
All iterates are projected into a Euclidean ball of radius $B$: $\|w_i^t\| \le B$ for all $i$ and $t$.

\paragraph{Assumption 5 (Network Topology) \cite{koloskova2019decentralized}}%
\label{assump:network}
The communication graph is connected and induces a symmetric, doubly-stochastic matrix $W$ with spectral norm gap $\sigma := \|W - \frac{1}{N} \mathbf{1}\mathbf{1}^\top\|_2 < 1$.

\paragraph{Assumption 6 (Smoothness of Alignment Loss)}%
\label{assump:bounded_alignment_gradient}
The auxiliary alignment loss $L_{\text{align}}(w)$ is $L_{\text{align}}$-smooth, i.e., $\|\nabla L_{\text{align}}(w)-\nabla L_{\text{align}}(v)\| \;\le\; L_{\text{align}}\|w-v\|, \quad \forall\, w,v\in\mathbb{R}^d,$
and its gradient is uniformly bounded: $ \|\nabla L_{\text{align}}(w)\|\le G_{\text{align}} $
for all $w$ in the feasible domain.

\paragraph{Theorem 1 (Convergence of DFedCAD)}%
\label{thm:dfedcad_main}
Under Assumptions 1–6, and with a sufficiently small fixed step size $\eta > 0$, the iterates of the DFedCAD algorithm satisfy:
\begin{equation}
\frac{1}{T} \sum_{t=0}^{T-1} \mathbb{E}\left[ \| \nabla F(\bar{w}^t) \|^2 \right]
\le \frac{2L}{\mu \eta T} \left( F(\bar{w}^0) - F^* \right) + \eta \cdot \mathcal{C},
\end{equation}
where $\bar{w}^t = \frac{1}{N} \sum_{i=1}^N w_i^t$ and $F^* = F(w^*)$. The constant $\mathcal{C}$ and all parameter conditions required for convergence are specified in Appendix E.

\begin{table*}[t]
\centering
\small
\setlength{\tabcolsep}{3pt}
\caption{Combined results on delayed clients across three benchmarks.
Missing entries are denoted by “–”. Best results in each column are in \textbf{bold}.}
\label{tab:main_results}
\begin{tabular}{llccc|ccc|cc|cc}
\toprule
\multirow{2}{*}{\textbf{Task}} & \multirow{2}{*}{\textbf{Method}} &
\multicolumn{3}{c|}{\textbf{Dirichlet $\alpha = 0.1$}} &
\multicolumn{3}{c|}{\textbf{Dirichlet $\alpha = 0.4$}} &
\multicolumn{2}{c|}{\textbf{Comm.\ Cost}} &
\multicolumn{2}{c}{\textbf{Comp.\ Cost}} \\
&  & Acc.\,$\uparrow$ & Var\,$\downarrow$ & $\pm$SD\,$\downarrow$ &
     Acc.\,$\uparrow$ & Var\,$\downarrow$ & $\pm$SD\,$\downarrow$ &
     (MB)$\downarrow$ & Red.\,\%$\uparrow$ &
     (FLOPs)$\downarrow$ & Red.\,\%$\uparrow$ \\
\midrule
\multirow{7}{*}{CIFAR-10}
 & DFedPGP   & 46.52\% & $2.8\times10^{-3}$ & 5.34\% & 56.94\% & $6.9\times10^{-3}$ & 8.28\% & 4.30 & 79.08 & 33.55\,B & $-0.15$ \\
 & QFedCG    & 49.52\% & $2.1\times10^{-3}$ & 4.59\% & 61.27\% & $1.3\times10^{-2}$ & 11.46\% & 8.14 & 60.39 & 33.51\,B & $-0.03$ \\
 & MTKD-RL   & 50.94\% & $2.3\times10^{-3}$ & 4.81\% & 61.77\% & $9.8\times10^{-3}$ & 9.91\% & – & – & 144.72\,B & $-332.00$ \\
 & ReT-FHD   & 46.55\% & $3.3\times10^{-3}$ & 5.75\% & 57.03\% & $1.4\times10^{-2}$ & 11.90\% & – & – & 44.67\,B & $-33.34$ \\
 & DFedSAM   & 48.07\% & $2.9\times10^{-3}$ & 5.38\% & 49.51\% & $2.2\times10^{-3}$ & 4.68\% & – & – & 66.92\,B & $-99.76$ \\
 & DFedAvg   & 58.80\% & $2.5\times10^{-3}$ & 5.02\% & 56.98\% & $7.2\times10^{-3}$ & 8.46\% & 20.55 & 0.00 & 33.50\,B & 0.00 \\
 & \textbf{DFedCAD} & \textbf{61.59\%} & $2.3\times10^{-3}$ & 4.76\% &
                       \textbf{62.09\%} & $7.9\times10^{-3}$ & 8.86\% &
                       \textbf{2.60} & 87.35 & \textbf{19.46\,B} & 41.91 \\
\midrule
\multirow{7}{*}{CIFAR-100}
 & DFedPGP   & 22.15\% & $1.1\times10^{-3}$ & 3.29\% & 29.95\% & $3.0\times10^{-3}$ & 5.44\% & 894.89 & 0.46 & 137.13\,B & $-0.02$ \\
 & QFedCG    & 21.99\% & $3.9\times10^{-4}$ & 1.98\% & 31.88\% & $3.0\times10^{-3}$ & 5.50\% & 408.42 & 54.57 & 137.20\,B & $-0.07$ \\
 & MTKD-RL   & 29.74\% & $2.2\times10^{-3}$ & 4.68\% & 37.03\% & $4.4\times10^{-3}$ & 6.63\% & – & – & 597.60\,B & $-335.89$ \\
 & ReT-FHD   & 14.31\% & $2.8\times10^{-4}$ & 1.67\% & 21.72\% & $1.7\times10^{-3}$ & 2.66\% & – & – & 182.34\,B & $-33.00$ \\
 & DFedSAM   & 37.85\% & $2.7\times10^{-3}$ & 5.19\% & 41.81\% & $1.1\times10^{-3}$ & 3.30\% & – & – & 274.10\,B & $-99.93$ \\
 & DFedAvg   & 35.12\% & $1.2\times10^{-3}$ & 3.41\% & 34.84\% & $7.1\times10^{-4}$ & 2.66\% & 899.00 & 0.00 & 137.10\,B & 0.00 \\
 & \textbf{DFedCAD} & \textbf{39.00\%} & $2.3\times10^{-3}$ & 4.78\% &
                       \textbf{42.27\%} & $3.9\times10^{-3}$ & 6.23\% &
                       \textbf{125.02} & 86.09 & \textbf{92.00\,B} & 32.90 \\
\midrule
\multirow{7}{*}{Tiny-ImageNet}
 & DFedPGP   & 24.19\% & $2.8\times10^{-4}$ & 1.67\% & 33.82\% & $4.1\times10^{-4}$ & 2.01\% & 894.89 & 0.46 & 1.05\,T & 0.00 \\
 & QFedCG    & 21.06\% & $2.8\times10^{-5}$ & 0.53\% & 31.81\% & $2.0\times10^{-4}$ & 1.41\% & 408.42 & 54.57 & 1.05\,T & 0.00 \\
 & MTKD-RL   & 22.84\% & $5.4\times10^{-5}$ & 0.73\% & 34.25\% & $8.33\times10^{-4}$ & 2.89\% & – & – & 4.66\,T & $-343.81$ \\
 & ReT-FHD   & 6.87\% & $2.0\times10^{-4}$ & 1.41\% & 11.22\% & $9.2\times10^{-4}$ & 3.04\% & – & – & 1.41\,T & $-34.29$ \\
 & DFedSAM   & 43.29\% & $8.7\times10^{-4}$ & 2.95\% & 48.48\% & $6.0\times10^{-4}$ & 2.44\% & – & – & 2.11\,T & $-100.95$ \\
 & DFedAvg   & 42.49\% & $1.1\times10^{-3}$ & 3.36\% & 44.14\% & $4.0\times10^{-4}$ & 1.99\% & 899.00 & 0.00 & 1.05\,T & 0.00 \\
 & \textbf{DFedCAD} & \textbf{46.57\%} & $4.2\times10^{-4}$ & 2.04\% &
                       \textbf{52.95\%} & $3.2\times10^{-4}$ & 1.79\% &
                       \textbf{125.02} & 86.09 & \textbf{0.86\,T} & 18.10 \\
\bottomrule
\end{tabular}
\end{table*}

\section{Experiment}
We conduct a series of experiments to rigorously evaluate our proposed DFedCAD framework. Our primary objective is to validate its core capability: accelerating the knowledge adaptation of late-joining clients in decentralized networks characterized by client delays and data heterogeneity. To this end, we will analyze the adaptation speed and final performance of these delayed clients. We aim to demonstrate that this rapid adaptation is achieved while simultaneously maintaining state-of-the-art model accuracy and achieving significant communication efficiency, thereby showcasing the comprehensive advantages of our method. Our source code is available at \footnote{https://anonymous.4open.science/r/DecentralizedFedLab-BB42}. A more detailed experimental configuration can be found in Appendix F.1.

\subsection{Experimental Setup}
\subsubsection{Datasets and Models}
We evaluate our method on three widely-used benchmark datasets: CIFAR-10, CIFAR-100, and Tiny-ImageNet \cite{krizhevsky2009learning, le2015tiny}. For the model architecture, we employ LeNet \cite{lecun1998gradient} for the CIFAR-10 dataset. For the more complex CIFAR-100 and Tiny-ImageNet datasets, which feature a larger number of classes and higher image resolution, we utilize the ResNet-18 \cite{he2016deep} architecture to ensure sufficient model capacity.

\subsubsection{Data Partitioning and Heterogeneity}
To simulate the statistically heterogeneous nature in real-world federated learning scenarios, we partition each dataset with a Dirichlet distribution using two concentration parameters, $\alpha=0.1$ and $\alpha=0.4$. 
Specifically, the CIFAR-10 dataset is distributed among 100 clients, while the more complex CIFAR-100 and Tiny-ImageNet datasets are each partitioned across 50 clients. For robust evaluation, each client's local data is further split into training and testing sets to ensure that model performance is assessed on unseen data from the same local distribution.

\subsubsection{Baseline Methods}
To comprehensively evaluate the performance of DFedCAD, we select a diverse set of baselines representing the state-of-the-art across different facets of federated learning. We include DFedAvg, a vanilla decentralized averaging method, to serve as a fundamental performance benchmark. To compare against methods addressing specific DFL challenges, we include DFedSAM \cite{shi2023improving} for its focus on mitigating statistical heterogeneity and DFedPGP \cite{liu2024decentralized} as a leading communication-efficient DFL framework. Furthermore, given our method's core is a novel distillation strategy, we make critical comparisons with ReT-FHD  \cite{qirethinking}, a knowledge distillation method for DFL, and MTKD-RL \cite{yang2025multi}, a state-of-the-art multi-teacher KD framework that we extend to the decentralized paradigm. Finally, to contextualize our results within the broader FL landscape, we include QFedCG \cite{xu2023federated}, a prominent communication-efficient method from the centralized domain.

\subsubsection{Delayed Client Scenario and Metrics}
To rigorously evaluate the adaptation capabilities of our method, we design a specific delayed client scenario. For each dataset, we first designate a fixed subset of 10\% of the clients to serve as the pool of potential delayed participants. This corresponds to 10 clients for CIFAR-10 and 5 clients for both CIFAR-100 and Tiny-ImageNet. The core of our evaluation consists of a series of independent experimental runs. In each run, a single, unique client from this pre-selected pool is activated to join the training process at a fixed late stage—specifically at communication round 25—while all other clients participate from the beginning. This process is repeated for every client in the delayed pool, ensuring that each one's unique data distribution is tested. The final reported results are averaged across these independent runs to provide a robust and unbiased assessment of performance. We measure effectiveness by the delayed client's per-round average Top-1 accuracy, to reflect adaptation speed, and its final average Top-1 accuracy, indicating peak performance. We quantify efficiency by the communication overhead (MB per round) and computational cost (FLOPs per round).

\subsection{Main Results on Delayed Client Adaptation}
The primary results of our experiments, focusing on the final performance of delayed clients, are summarized in Table~\ref{tab:main_results}. This table reports the maximum average Top-1 accuracy, along with communication and computation overheads, of delayed clients across three datasets under two levels of data heterogeneity. To ensure a fair comparison, methods not explicitly designed for model communication reduction, such as MTKD-RL and DFedSAM, are excluded from the communication analysis.

The results clearly demonstrate the superiority of DFedCAD. Across all datasets and Dirichlet imbalance levels, it consistently outperforms baseline methods, showcasing strong adaptability for clients joining late in training.


In particular, the performance advantage of DFedCAD is particularly pronounced in the more challenging, highly heterogeneous setting ($\alpha=0.1$). For instance, on the complex Tiny-ImageNet dataset, DFedCAD achieves an accuracy of 46.57\%. Similarly, on CIFAR-100 under the same heterogeneity, DFedCAD (39.00\%) shows a clear improvement over all other approaches. Even in moderately heterogeneous environments ($\alpha=0.4$), DFedCAD maintains its leading position, achieving an accuracy of 52.95\% on Tiny-ImageNet.

Beyond accuracy, DFedCAD also delivers substantial communication and computation savings. On CIFAR-10, it reduces communication by 87.35\% and computation by 41.91\% compared to DFedAvg. Similar trends hold across other datasets, underscoring DFedCAD’s effectiveness. A detailed analysis is provided in Appendices F.3 and F.4.


\subsection{Supplementary Experiments}
To further validate DFedCAD, we present additional experiments in the Appendix F. Specifically, we perform an ablation study examining the impact of the centroid-aligned distillation module in the Appendix F.2. The results clearly demonstrate that structural alignment significantly improves delayed-client adaptation, underscoring the necessity of the proposed distillation strategy.
Beyond the main results, Appendices F.3 and F.4 provide a deeper breakdown of communication and computation costs, complementing the summary in the main text.
Finally, we provide learning curves in Appendix F.5, demonstrating the convergence behavior of the model under different settings.

\section{Conclusion}
We propose DFedCAD, a novel decentralized federated learning framework for rapid adaptation of delayed clients under communication constraints. DFedCAD leverages centroid-aligned distillation, combining Weight Clustering Pruning, centroid-distribution distance, and differentiable $k$-means alignment. Extensive experiments demonstrate that DFedCAD consistently outperforms state-of-the-art methods in accuracy and efficiency, significantly reducing communication overhead. Our work offers a 
feasible solution for decentralized learning in dynamic real-world scenarios.

\bibliography{aaai2026}
\end{document}